\documentclass[lettersize,journal]{IEEEtran}
\usepackage{amsmath,amsfonts}
\usepackage{algorithmic}
\usepackage{algorithm}
\usepackage{array}
\usepackage[caption=false,font=normalsize,labelfont=sf,textfont=sf]{subfig}
\usepackage{textcomp}
\usepackage{stfloats}
\usepackage{url}
\usepackage{verbatim}
\usepackage{graphicx}
\usepackage{cite}
\usepackage{multirow}
\usepackage{arydshln}
\usepackage[colorlinks,linkcolor=blue]{hyperref}

\begin{document}

\title{$\textup{C}^2$Former: Calibrated and Complementary Transformer for  RGB-Infrared Object Detection}

\author{Maoxun Yuan and Xingxing Wei$^\ddagger$
\thanks{Maoxun Yuan is with the Beijing Key Laboratory of Digital Media, School of Computer Science and Engineering, Beihang University, Beijing
100191, China (email: yuanmaoxun@buaa.edu.cn).}

\thanks{Xingxing Wei is with Institute of Artificial Intelligence, Hangzhou Innovation Institute, Beihang University, Beijing, 100191, China (email: xxwei@buaa.edu.cn). }
\thanks{$\ddagger$ indicates the corresponding author.}
}

\markboth{IEEE TRANSACTIONS ON GEOSCIENCE AND REMOTE SENSING}%
{Yuan \MakeLowercase{\textit{et al.}}: Calibrated and Complementary Transformer for RGB-Infrared Object Detection}


\maketitle

\begin{abstract}
    Object detection on visible (RGB) and infrared (IR) images, as an emerging solution to facilitate robust detection for around-the-clock applications, has received extensive attention in recent years. With the help of IR images, object detectors have been more reliable and robust in practical applications by using RGB-IR combined information. However, existing methods still suffer from modality miscalibration and fusion imprecision problems. Since transformer has the powerful capability to model the pairwise correlations between different features, in this paper, we propose a novel \textbf{C}alibrated and \textbf{C}omplementary Trans\textbf{former} called $\textbf{C}^2\textbf{Former}$ to address these two problems simultaneously. In $\textbf{C}^2\textbf{Former}$, we design an Inter-modality Cross-Attention (ICA) module to obtain the calibrated and complementary features by learning the cross-attention relationship between the RGB and IR modality. To reduce the computational cost caused by computing the global attention in ICA, an Adaptive Feature Sampling (AFS) module is introduced to decrease the dimension of feature maps.  Because $\textbf{C}^2\textbf{Former}$ performs in the feature domain, it can be embedded into existed RGB-IR object detectors via the backbone network. Thus, one single-stage and one two-stage object detector both incorporating our $\textbf{C}^2\textbf{Former}$ are constructed to evaluate its effectiveness and versatility. With extensive experiments on the DroneVehicle and KAIST RGB-IR datasets, we verify that our method can fully utilize the RGB-IR complementary information and achieve robust detection results. 
    The code is available at \href{https://github.com/yuanmaoxun/C2Former.git}{https://github.com/yuanmaoxun/C2Former.git}.
\end{abstract}

\begin{IEEEkeywords}
RGB-infrared object detection, modality calibration, complementary fusion, multispectral object detection.
\end{IEEEkeywords}

\section{Introduction}
Object detection, as a core technology in the computer vision field, has been used in various practical applications, such as autonomous driving, video surveillance and robotics, etc. With the development of convolutional neural network (CNN), the past few years have witnessed significant progress in object detectors  \cite{ren2015faster,redmon2016you, lin2017focal, duan2019centernet, bo2021ship, xie2021weakly, xie2020ship, wang2020ship}. Since these detectors are mainly designed for visible (RGB) images, they cannot cope with the challenges in bad illumination (nighttime) or diverse weather conditions (rain, dust and fog) \cite{kieu2019domain, li2018multispectral,wei2023unified}. For these reasons, detectors exploiting infrared images have been widely adopted as an additional modality for robust object detection \cite{herrmann2018cnn, devaguptapu2019borrow, kieu2020task}. To achieve better performance in the full-time detection task, a growing number of works \cite{zhang2019cross, guan2019fusion, xie2023co,li2019illumination} have been investigated to obtain rich features of objects by integrating  visible (RGB) cues and infrared (IR) spectra. 

\begin{figure}[!t]
	\begin{center}
	\includegraphics[scale=0.45]{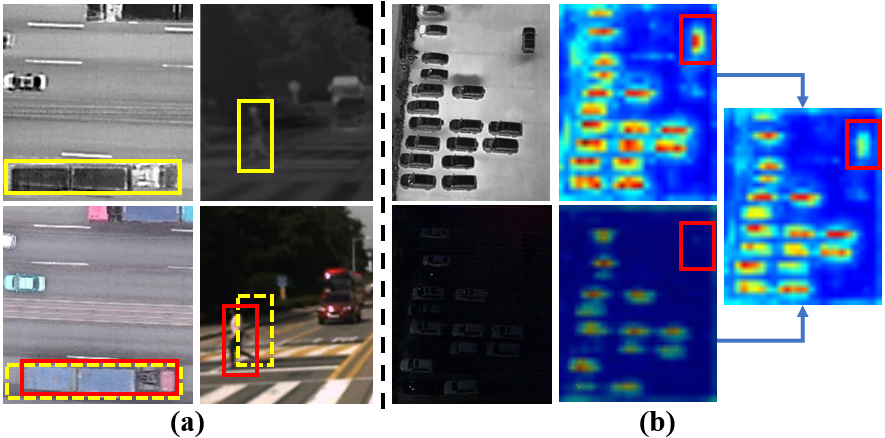}
	\end{center}
    \caption{Two major problems of RGB-IR object detection. (a) An example of modality miscalibration between RGB and Infrared modalities.  The yellow and red boxes represent annotations of same objects in the IR images and the RGB images, respectively. (b) An example of fusion imprecision output by MBNet \cite{zhou2020improving}. We see that the fusion features in the red boxes are even worse than the infrared feature, which shows the difficulty in feature fusion.}
    \label{fig1_a}
\end{figure}

However, there are still two major problems in RGB-IR object detection \cite{kim2021uncertainty,zhou2020improving}. (1) One is \textbf{modality miscalibration} as shown in Figure~\ref{fig1_a}(a). Existing methods \cite{li2019rgb,liu2021cross,guan2019fusion,liu2021learning,li2023robust} usually assume that RGB-IR image pairs are perfectly geometrically aligned, and they directly perform the multimodal fusion methods.  Actually, after the image registration algorithm, the paired images are just weakly aligned \cite{zhang2021weakly,yuan2022translation}. This is because the RGB-IR image pairs are captured by different sensors with different field-of-views (FoVs) and imaging time, the positions of moving objects may be out of synchronization. 
(2) Another problem is \textbf{fusion imprecision} as shown in Figure~\ref{fig1_a}(b). In IR modality, the thermal radiation of the vehicle can be important hints for the detector, but none of the RGB images has such cues in poor illumination. The RGB and IR modality features are diverse in terms of object color, texture, and properties, as they are captured in different spectral bands, which brings a big challenge to the fusion strategy.
In fact, these two problems are tightly coupled, the misaligned features will directly lead to the imprecision fusion. Thus, these two problems need to be bound together and solved simultaneously.

Recently, some approaches have been introduced to solve these problems in RGB-T detection task. Zhang et al. \cite{zhang2021weakly} and Yuan et al. \cite{yuan2022translation} solve the modality miscalibration problem by predicting the proposal offset between two modalities in the RoI head. Although these instance-level prediction solves the object misalignment problem to some extent, they cannot guarantee that the features inside two modality proposals are perfectly aligned.  Besides, some cross-modality fusion strategies \cite{li2018multispectral, xia2022dml, zhang2019cross} are also conducted to address the fusion imprecision problem through simple feature concatenation or illumination-aware fusion.  However, the inherent complementary information between RGB-IR modality features is still not exploited. More importantly, the above methods do not consider correlating the problems of modality miscalibration and fusion imprecision, but mainly focus on one of them or solve them separately, which will affect the detection performance.

\begin{figure}[t]
\centering
\includegraphics[scale=0.5]{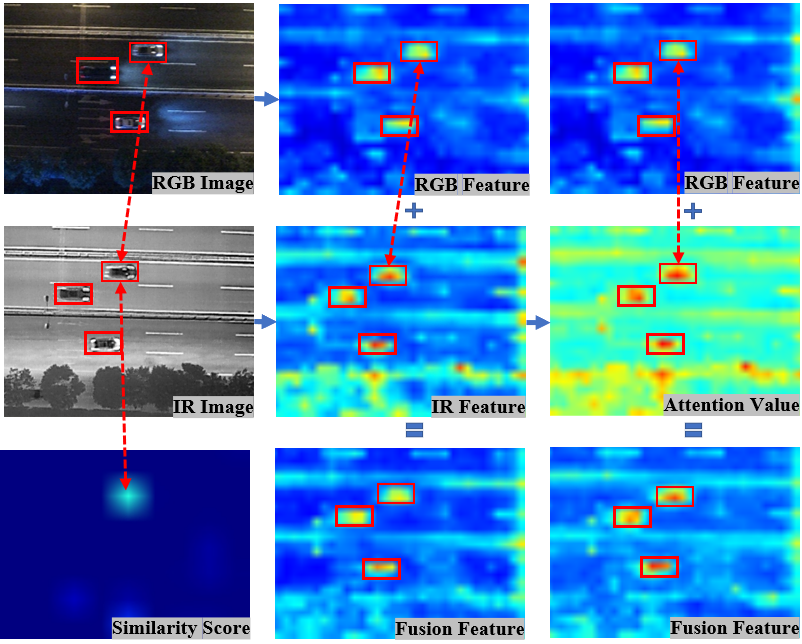}
\caption{An illustration result of ICA module. We see  the attention values are perfectly aligned with the referenced RGB features and enhance the objects' regions compared with the original IR features, which are more complementary to RGB features. }
\label{fig1_b}
\end{figure}

In this paper, referring to the self-attention mechanism in the transformer \cite{vaswani2017attention}, we formulate the aforementioned two problems into transformer framework and propose a novel \textbf{C}alibrated and \textbf{C}omplementary Trans\textbf{Former}-based fusion method called \textbf{$\textup{C}^2$Former} to these two problems simultaneously. Specifically, for each feature point in one modality (Query vector), we first compute the similarity score with all the feature points in the other modality (Key vectors), and then obtain the attention values via the weighted sum versus the features in the other modality (Value vectors). These attention values have two properties: (1) they incorporate the global context information from the other modality, therefore, can be regarded as a better complementary feature than the single local feature point to the original modality. Thus, we can simply concatenate these attention values with the original feature point to conduct an effective feature fusion. (2) The most relevant feature points in the other modality contribute the most to the attention values, which implies that the calibration is automatically achieved thanks to the similarity score. In this way, we simultaneously perform the calibration and fusion operations in a unified framework. Based on this idea, we design an Inter-modality Cross-Attention (ICA) module. An illustration result of the ICA module is given in Figure~\ref{fig1_b}.

Directly computing attention across global feature maps has high computational costs. For that, it needs a feature sampling to reduce the dimension of feature maps. However, because of the miscalibration, an  uniform sampling strategy for both modalities may omit meaningful features. To tackle this issue, we present an Adaptive Feature Sampling (AFS) strategy. In detail,  we predict an offset versus the foreground object across modalities through a deviation network, and then generate the corresponding sampling strategy according to this offset for different modalities. In this way, we can ensure the integrity of important regions in the sampled feature maps.  In actual usage, the given feature maps are first down-sampled by AFS module, and then processed by ICA module to obtain the complementary features, which are finally concatenated with the given feature maps to obtain the fusion features. 

In summary, our contributions are listed as follows:

\begin{itemize}
    \item[$\bullet$] We propose a novel transformer-based network called $\textup{C}^2$Former to simultaneously solve the modality miscalibration and fusion imprecison problems specific to the RGB-IR object detection task.
    \item[$\bullet$] We propose an ICA module to simultaneously align and fuse the RGB-IR features, and present an AFS module to reduce the computational complexity in ICA. These two modules interact with each other and can be combined in an end-to-end manner.
    \item[$\bullet$] To evaluate the effectiveness and flexibility of our $\textup{C}^2$Former, we construct two $\textup{C}^2$Former-based detectors for RGB-IR object detection. Experiments on the challenging RGB-IR objects detection datasets show that the detectors achieve the SOTA performance. 
\end{itemize}

The remainder of the article is organized as follows. In the next section, we review the scientific literature related to our proposed approach. Section \ref{Method} analyzes the problem and describes our approach. Experimental results are presented in Section \ref{experiments}, along with the comparison with state-of-the-art methods, and in Section \ref{conclusion} we conclude with a discussion of our contribution.

\section{Related Work} \label{related}
\subsection{Multi-modal Transformers}
Since the transformer model \cite{vaswani2017attention} has shown its powerful ability to capture long-range dependencies, multi-modal transformers \cite{liu2021tritransnet,liu2022learning,yuan2024improving,wang2022bridged,xiao2021fetnet,Shi_2023_CVPR,xia2018ViTCoMer} have also been explored. To fuse different modality features, CMSA \cite{ye2019cross} feeds the words and images into the transformer to construct a joint multimodal feature representation for referring image segmentation and BrT \cite{wang2022bridged} uses a transformer to allow interactive learning by adding a point-to-patch projection between images and point clouds. Similarly, Liu et al. \cite{liu2021tritransnet} and Xiao et al. \cite{xiao2021fetnet} design a triplet transformer and feature exchange transformer respectively to detect the salient objects in RGB-D images. However, these methods perform feature fusion by directly concatenating or summing the multi-modalities to feed the transformer. They do not consider the feature misalignment problem, which will lead to a limited representation ability of the fused features.

Different from above approaches, Pixel-BERT \cite{huang2020pixel} considers the modality misalignment problem and aligns vision and language semantics at the pixel and text level using self-supervised learning. Besides, Wang et al. \cite{wang2021dig} leverage the hierarchical alignment to explore the detailed and diverse characteristics in text-video retrieval task.  Recently, Wang et al. \cite{wang2022multimodal} aggregate RGB and depth cues through projecting alignment features from other modalities at the token level.  Even though these methods consider the feature misalignment problem, the modality complementarity is not further exploited in the fusion process. 

Above approaches attempt to utilize the transformer to address multi-modal problems. However, compared with these multi-modal tasks (words-images, texts-videos and point clouds-images), our $\textup{C}^2$Former is designed specific to the RGB-T object detection task and we are the first to apply transformer to this task. Since the problems of modality miscalibration and fusion imprecision are unique and intractable in RGB-T image pairs, we are inspired by the core idea of the self-attention mechanism and formulate these two problems into transformer to establish their correlation. Thus, these two problems are combined and solved simultaneously, which is of great significance in RGB-T object detection task.


\subsection{Aerial Object Detection}
Aerial object detection refers to using rotated bounding box to represent building detectors, which can provide more accurate location of objects. Xia et al.  \cite{xia2018dota} construct a large-scale object detection benchmark with oriented annotations, named DOTA. Existing methods of building detectors based on horizontal bounding boxes are plagued by multiple objects of interest in one anchor. Therefore, many researches about aerial object detection \cite{ma2018arbitrary,jiang2017r2cnn,ding2019learning,yan2022antijamming,shi2020orientation,zhu2020adaptive} have been presented to alleviate this problem. Ma et al. \cite{ma2018arbitrary} construct R-RPN to detect aerial objects using rotated proposals. Jiang et al. present $\rm{R^2}$CNN \cite{jiang2017r2cnn}, which has a horizontal region of interest (RoI) that is leveraged to predict both horizontal and rotated boxes. To avoid a large number of anchors, RoI Transformer \cite{ding2019learning} is proposed to transform the horizontal RoI in $\rm{R^2}$CNN into a rotated one.  To speed up the inference time of detectors, some works \cite{yang2021r3det,han2021align,cheng2022anchor,ming2021cfc} explore one-stage and anchor-free oriented object detectors. $\rm R^3$Det  \cite{yang2021r3det} and $\rm S^2$A-Net \cite{han2021align} adopt alignments between horizontal receptive fields and rotated anchors to get a better feature representation.  Recently, GWD \cite{yang2021rethinking} and KLD \cite{yang2021learning} are proposed to use the Gaussian Wasserstein distance and KL divergence to optimize the localization of the bounding boxes, respectively. 

In recent years, utilizing RGB and infrared images for aerial object detection has also received widespread attention. Sun et al. \cite{sun2022drone} provide a benchmark dataset called DroneVehicle for RGB-IR vehicle detection. On the basis of this dataset, a novel and lightweight detector named CMAFF \cite{qingyun2022cross} has been proposed to perform multispectral feature fusion. To further improve vehicle detection performance at nighttime, Gao et al. \cite{gao2022gf} present a vehicle detection-driven RGB and infrared image fusion method called GF-detection. Furthermore, Yuan et al. \cite{yuan2022translation} observe the weak misalignment problem in DroneVehicle and introduce a TSRA module to calibrate two modality features. Recently, the auxiliary super-resolution branch proposed by \cite{zhang2023superyolo} has been introduced into RGB-IR objecrt detection, which can effectively distinguish small objects.
In this paper, we also build the two-stream aerial object detector incorporating our proposed $\textup{C}^2$Former to perform complementary feature fusion.


\subsection{Multispectral Pedestrian Detection}
Multispectral pedestrian detection is an important research area for pedestrian detection and has achieved remarkable results. Since the KAIST dataset \cite{hwang2015multispectral} was released, more and more studies have been proposed to improve detector performance by utilizing aligned RGB and IR images. Wagner et al. \cite{wagner2016multispectral} propose the first fusion architecture, which uses multiple modality images for fusion to improve the reliability of pedestrian detection. \cite{li2019illumination} and \cite{guan2019fusion} propose an illumination-aware fusion architecture to fuse RGB-IR features adaptively. 
Thus, Fusion CSPNet \cite{wolpert2020anchor} is presented to fuse features from different modalities for small-scale pedestrian and partially occluded instances.
Besides, Zhou et al. \cite{zhou2020improving} address the modality imbalance problem by designing a feature fusion module called DMAF. To solve the weakly misalignment problem between RGB-IR objects, ARCNN \cite{zhang2021weakly} adds an extra RFA module to the detection head to predict the offset between RGB-IR objects. However, the generalization of RFA module is poor, which can only be applied into the two-stage detectors. Therefore, we propose a new feature alignment module utilizing the cross-attention mechanism, which can be used in various existing detectors.

\section{Proposed Method} \label{Method}
In this section, we first introduce our proposed ICA module in Section \ref{ica}. To reduce its computational complexity, we propose an
AFS module in Section \ref{afs}. Finally, we put everything together into a description of the proposed detection framework in Section \ref{overall-architecture}.
\subsection{Inter-modality Cross-Attention (ICA)} \label{ica}
\begin{figure*}[!t]
	\begin{center}
	\includegraphics[width=2.0\columnwidth]{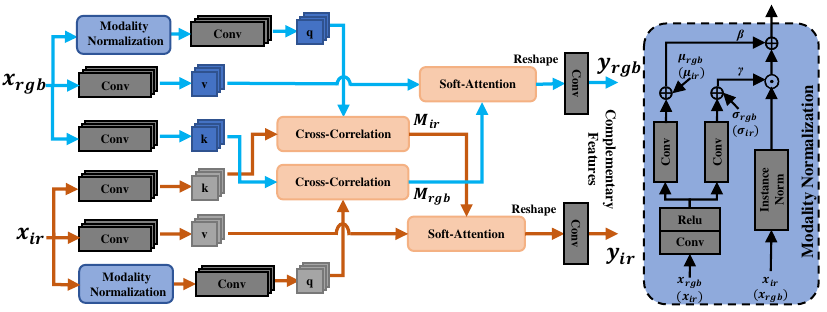}
	\end{center}
    \caption{The process of ICA module. $\bigoplus$ and $\bigodot$ indicate the addition and dot product operations respectively. In Modality Normalization, we first normalize the feature distribution of one modality and then inject the mean and variance predicted by the CNN network into the normalized feature distribution to achieve the transformation of the feature distribution.}
	\label{fig3}
\end{figure*}

Inter-modality Cross-Attention aims to identify aligned and complementary features, which will be further utilized to enhance feature representations for the backbone network. To facilitate ICA module (shown in Figure~\ref{fig3}), cross-attention mechanism is utilized to establish the cross-correlations between RGB-IR features. Specifically, we follow self-attention mechanism to derive a set of $N$ descriptors for each modality feature in $q$,  $k$ and $v$ by using convolution layers. Next, we compute similarity matrices to find aligned features through cross-correlation alignment and obtain complementary features for each modality after soft-attention fusion. To reduce the variance between different modality descriptors, we also design a Modality Normalization process. Finally, we reshape the output feature from soft-attention and adopt convolution layers to convert them back to the original feature space. The designed ICA module greatly helps the $\textup{C}^2$Former in finding cross-modal correlations between RGB and IR features.

\noindent\textbf{Acquiring Feature Descriptors.}
Assume that the input RGB-IR features of the ICA module are $x_{rgb} \in \mathbb{R}^{C \times H \times W}$ and $x_{ir} \in \mathbb{R}^{C \times H \times W}$, we first use the the input features to generate query $q$, key $k$ and value $v$ descriptors, respectively.
Specifically, we employ convolution layers for each modality to generate descriptors and reshape them into 2D tensors. The generation process is defined:
\begin{equation}
    \begin{aligned} 
    &q_{r g b}=\Gamma(W_{r g b}^q * x_{r g b}), \quad q_{i r}=\Gamma(W_{i r}^q * x_{i r}),  \\
    &k_{r g b}=\Gamma(W_{r g b}^k * x_{r g b}), \quad k_{i r}=\Gamma(W_{i r}^k * x_{i r}),  \\
    &v_{r g b}=\Gamma(W_{r g b}^v * x_{r g b}), \quad v_{i r}=\Gamma(W_{i r}^v * x_{i r}), 
    \end{aligned}
    \label{eq:eq1}
\end{equation}
where $W^{q}$, $W^{k}$ and $W^{v}$ are the convolution with kernel size $1\times1$ associated with query, key and value respectively. $q \in \mathbb{R}^{HW \times C}$, $k \in \mathbb{R}^{HW \times C}$ and $v \in \mathbb{R}^{HW \times C}$ are the final feature descriptors after a reshape operation $\Gamma(\cdot)$.

\begin{figure}[!t]
	\begin{center}
	\includegraphics[scale=0.33]{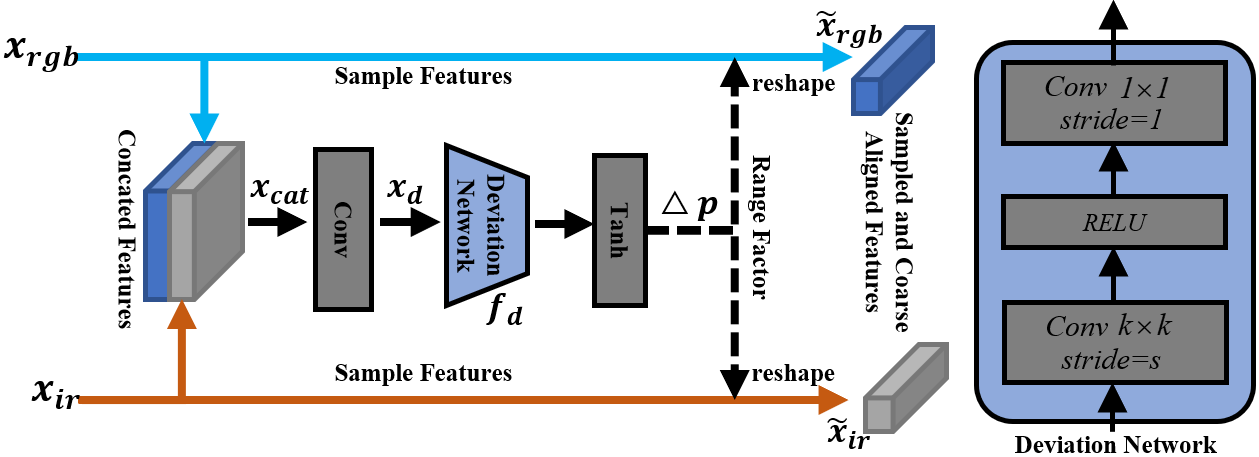}
	\end{center}
    \caption{The structure of AFS module, which is used to obtain sampled and coarse-aligned features.}
	\label{fig2}
\end{figure}

\noindent\textbf{Modality Normalization.} 
Since RGB and IR images are captured at different spectral bands, there is modality discrepancy between two images. Thus, directly using image feature from one modality to compute the similarity score with image feature from the other modality is not optimal. Inspired by \cite{kim2016deeply}, we propose the Modality Normalization to transform the image feature from one modality into the distribution of the other modality. The process of Modality Normalization is given on the right of Figure~\ref{fig3}, which considers the mean and variance from the other modality.

The process of transforming two modalities features is the same. Here, we take the process of IR modality features as an example to perform Modality Normalization. We first apply an instance normalization \cite{ulyanov2016instance} to the IR feature $x_{ir}$:
\begin{equation}
    \tilde{x}_{ir} = \frac{x_{ir}-\mu_{ir}}{\sigma_{ir}},
\end{equation}
where $\sigma_{ir}$ and $\mu_{ir}$ are the mean and standard deviation tensor of $I_{ir}$. We also calculate the standard deviation $\mu_{rgb}$ and mean deviation $\sigma_{rgb}$ of $x_{rgb}$, which will be used to remap the $x_{ir}$ distribution. Then three convolution layers are used to predict two learnable parameter tensors $\beta_{ir}$ and $\gamma_{ir}$. After that, we add $\sigma_{rgb}$ and $\mu_{rgb}$ to the $\gamma_{ir}$ and $\beta_{ir}$ respectively:
\begin{equation}
    \begin{aligned}
    &\beta_{ir}=W_b * \operatorname{relu}\left(W_d * x_{r g b}\right)+\mu_{r g b}, \\
    &\gamma_{ir}=W_g * \operatorname{relu}\left(W_d * x_{r g b}\right)+\sigma_{r g b},
    \end{aligned}
\end{equation}
where $W_b$, $W_g$ and $W_d$ are the convolution with kernel size $3\times3$. Finally, we obtain the transformed IR features by integrating $\gamma_{ir}$ and $\beta_{ir}$ into the normalized IR feature $\tilde{I}_{i r}$. Similarly, the transformed RGB features can also be obtained:
\begin{equation}
    \begin{aligned}
        &\tilde{x}_{i r} = \tilde{x}_{i r} \cdot\gamma_{ir}+\beta_{ir}, \\
        &\tilde{x}_{rgb} = \tilde{x}_{rgb} \cdot\gamma_{rgb}+\beta_{rgb}.
    \end{aligned}
\end{equation}
Thus, the final generation process of query ($q_{rgb}$ and $q_{ir}$) descriptors in Equation(\ref{eq:eq1}) are modified as follows:
\begin{equation}
    \begin{aligned}
        &q_{r g b}=W_{r g b}^q * \tilde{x}_{r g b}, \\
        &q_{i r}=W_{i r}^q * \tilde{x}_{i r}.
    \end{aligned}
\end{equation}

\noindent\textbf{Cross-Correlation Alignment.}
To calibrate the features between RGB and IR modalities, we compute attention maps between the modality features relative to each other. we first perform the inner product operation between two modality features through matrix multiplication. Then, we normalize the result to obtain the similarity scores. After that, the most relevant feature points (aligned feature) in the other modality have the highest scores. Through above process, we construct the similarity matrices $\tilde{M} \in \mathbb{R}^{HW \times HW }$ of each modality. 
Specifically, we apply matrix multiplication between query (\emph{e.g.} $q_{rgb}$) and key (\emph{e.g.} $k_{ir}$) for different modality descriptors. The RGB and IR matrices can be formulated as follows:
\begin{equation}
    \begin{aligned}
        &\tilde{M}_{r g b}=\operatorname{MatMul}\left(q_{i r}, k_{r g b}^{T}\right) / \sqrt{d}, \\
        &\tilde{M}_{i r}=\operatorname{MatMul}\left(q_{r g b}, k_{i r}^{T}\right) / \sqrt{d},
    \end{aligned}
\end{equation}
where $\operatorname{MatMul}$ denotes the matrix multiplication, $T$ represents the transpose operation of the matrix and $\sqrt{d}$ is a scaling parameter. We then employ the softmax layers to obtain the column-normalized (sums to 1) cross-correlation matrices $M_{rgb}$ and $M_{ir}$. Formally, we can define this process as follows:
\begin{equation}
    \begin{aligned}
        &M_{rgb}=\operatorname{Softmax}\left(\tilde{M}_{rgb}\right), \\
        &M_{i r}=\operatorname{Softmax}\left(\tilde{M}_{i r}\right),
    \end{aligned}
\end{equation}
where $M_{rgb}$ and $M_{ir}$ are the column-normalized (sums to 1) similarity matrices of $N$ descriptors.

 \noindent\textbf{Soft-Attention Fusion.} With the similarity matrices $M$, we can dynamically complement features according to the weights and inject contextual feature information to make the complementary features more robust. In this process, the matrix multiplication is also utilized between similarity matrices $M$ and value descriptors:
\begin{equation}
    \begin{aligned}
        &\tilde{y}_{rgb}=\operatorname{MatMul}\left(M_{rgb}, v_{rgb}\right), \\
        &\tilde{y}_{i r}=\operatorname{MatMul}\left(M_{i r}, v_{i r}\right),
    \end{aligned}
    \label{eq:eq2}
\end{equation}
Then, we reshape the dimension of $\tilde{y}_{rgb}$ and $\tilde{y}_{ir}$ to the original feature format ($\mathbb{R}^{HW \times C} \rightarrow \mathbb{R}^{C \times H \times W}$) and adopt convolution layers to obtain advanced complementary feature representation from $\textup{C}^2$Former. The final output features $y_{r g b}$ and $y_{i r}$ can be formulated as follows:
\begin{equation}
    \begin{aligned}
        &y_{rgb}=W_m*\Gamma\left(\tilde{y}_{rgb}\right), \\
        &y_{i r}=W_n*\Gamma\left(\tilde{y}_{i r}\right),
    \end{aligned}
    \label{eq:eq3}
\end{equation}
where $W_m$ and $W_n$ denote the convolution layers of the respective modalities.

\begin{figure*}[!t]
	\begin{center}
	\includegraphics[scale=0.35]{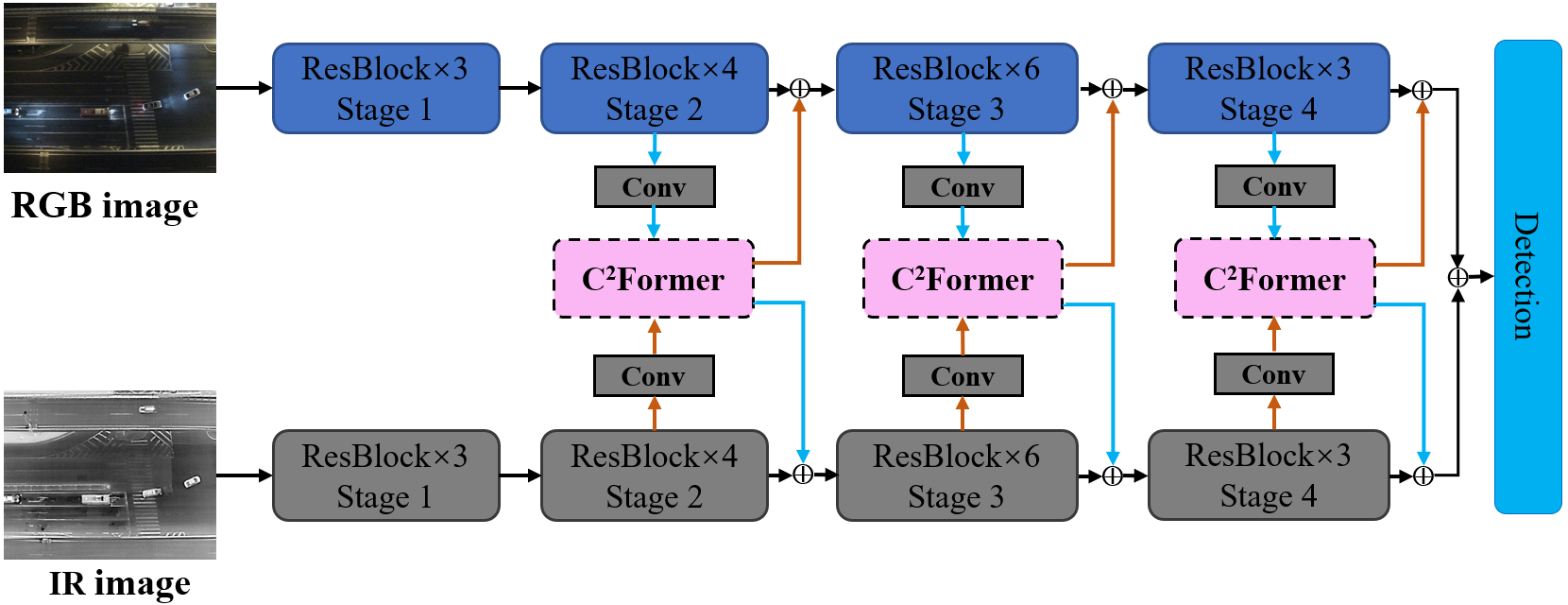}
	\end{center}
    \caption{The framework of the $\textup{C}^2$Former-based detectors. Our $\textup{C}^2$Former consists of ICA and AFS, where the input features are first reduced in feature dimension through the pre-module AFS and then output the aligned and fused features by the ICA module. The outputs of $\textup{C}^2$Former are added into opposite ResNet-50 backbone networks by using an addition operation.  For clear illustration, we do not show the FPN structure in this framework. Same as the last layer operation, we add the features output by each stage of the two modalities to obtain multi-scale features and finally input them into the FPN.}
	\label{fig4}
\end{figure*}

 \subsection{Adaptive Feature Sampling (AFS)} \label{afs}
Due to the high computational complexity of matrix multiplication in ICA module, we need to design a predecessor module for ICA to reduce the dimension of the input features. Thus, we introduce a novel feature sampling method called Adaptive Feature Sampling (AFS) to sample the features. To ensure that pairs of aligned features are not completely lost during the sampling process, we need to predict the coarse offset between two modality features and then sample the features according to the offset. Therefore, we design the AFS module based on deformable convolution \cite{dai2017deformable, zhu2019deformable}. The structure of AFS module is illustrated in Figure~\ref{fig2}. Firstly, we combine RGB-IR features and predict their feature offsets by using a deviation network (two convolution modules with a \emph{relu} activation). Based on the offsets, the coarse-aligned features are then dynamically sampled from the RGB and IR feature maps respectively by using the bilinear interpolation operation.

Specifically, the inputs of AFS are also $x_{rgb}$ and $x_{ir}$ since the AFS precedes the ICA module to sample the features. We firstly concat them $x_{cat} = \operatorname{Concat}\left(x_{r g b}, x_{i r}\right)$ and acquire a new feature map $x_{d} \in \mathbb{R}^{C \times H \times W} $ by utilizing a $1\times1$ convolution $W_{c}$ to reduce the channel dimension, $x_{d}=W_{c}*x_{cat}$. Note that we discard the batch dimension from our notations for simplicity. Then, we use a deviation network $f_{\text {d}}$ to generate the offsets $\Delta p \in \mathbb{R}^{H \times W \times 2}$ between two modalities. Since $\Delta p$ may fluctuate during training, we scale the range of $\Delta p$ by using an activation function \emph{tanh}. Therefore, we predict the RGB-IR feature offsets $\Delta p$, which can be formulated as:
\begin{equation}
    \Delta p=2 \cdot \tanh \left(f_{d}\left(x_{d}\right)\right).
\end{equation}
After that, we generate grids of reference points $p_{rgb}\in \mathbb{R}^{H_{s} \times W_{s} \times 2}$ and $p_{ir}\in \mathbb{R}^{H_{s} \times W_{s} \times 2}$ for the RGB and IR modalities, where $H_{s}=H/s$ and $W_{s}=W/s$ are downsampled from the input featrues' size by the stride factor $s$. Then, we normalize these points $p_{rgb}$ and $p_{ir}$ to the range $[-1,+1]$, where $(-1,-1)$ represents the top-left position and $(+1,+1)$ represents the bottom-right position. Finally, the sampled features from two modalities ($x_{r g b}$ and $x_{i r}$) are selected at the predicted locations ($p_{r g b}+\Delta p$ and $ p_{i r}$) by using the bilinear interpolation sampling function $\varphi(\cdot ; \cdot)$ as:
\begin{equation}
    \begin{aligned}
        &\bar{x}_{r g b}=\varphi\left(x_{r g b}; p_{r g b}+\Delta p\right), \\
        &\bar{x}_{i r}=\varphi\left(x_{i r}; p_{i r} \right).
    \end{aligned}
\end{equation}

Finally, we obtain the sampled features $\bar{x}_{r g b}  \in \mathbb{R}^{C \times H_{s} \times  W_{s}}$ and $\bar{x}_{ir}  \in \mathbb{R}^{C \times H_{s} \times  W_{s}}$, which are used as final input to the ICA module. Thus, the dimensions of the descriptors in Equation(\ref{eq:eq1}) and the outputs in Equation(\ref{eq:eq3}) become $\mathbb{R}^{H_{s}W_{s} \times C}$ and  $\mathbb{R}^{  C \times H_{s} \times W_{s}}$, respectively. To make the output ($y_{rgb}$ and $y_{ir}$) dimensions of ICA consistent with the input feature ($x_{rgb}$ and $x_{ir}$), $W_{m}$ and $W_{n}$ in Equation(\ref{eq:eq3}) become the $1\times1$ deconvolution \cite{zeiler2011adaptive} with upsample stride $s$.

\subsection{${C}^2$Former-based Object Detectors} \label{overall-architecture}
To evaluate our proposed $\textup{C}^2$Former, we design a framework of RGB-IR detectors incorporating $\textup{C}^2$Former as shown in Figure~\ref{fig4}. Since $\textup{C}^2$Former performs in the backbone network, different detection heads can be used after it. Respectively, we construct one-stage and two-stage detectors based on $\textup{S}^2$A-Net \cite{han2021align} and Cascade R-CNN \cite{cai2018cascade} for oriented and horizontal detection respectively. In this framework, ResNet-50 \cite{he2016deep} is utilized as baseline backbone and convolutional layers are used to reduce the channels of RGB-IR features. The input of each $\textup{C}^2$Former is the output of the corresponding stages of the two modalities. Then complementary features output by $\textup{C}^2$Former is added into opposite backbone networks to enhance the representation of modality features. To guarantee the information integrity of the different modality features, we finally feed the sum features of the two modalities into the object detection head.

\begin{table}[!t]\normalsize
\renewcommand{\arraystretch}{1.0}
\begin{center}
\caption{Ablation experiments for each component evaluated on DroneVehicle dataset.}
\centering
 \label{table3}
\setlength{\tabcolsep}{0.5mm}
\begin{tabular}{cc|ccccc:cc}
    \hline
    \textbf{ICA} & \textbf{AFS}  & \textbf{Car}  & \textbf{Truck} &\textbf{Fre.} & \textbf{Bus}  & \textbf{Van}  & \textbf{mAP}  & \textbf{FLOPs} \\ \hline
        &      & 90.0 & 64.5  & 61.7        & 88.2 & 53.2 & 71.5 & \textbf{97.8G} \\
     \checkmark   &     & 90.0 & 68.2  & \textbf{65.2}    & 88.8 & 56.8 & 73.8 & 139.8G \\
    \checkmark   & \checkmark        & \textbf{90.2} & \textbf{68.3}  & 64.4        & \textbf{89.8} & \textbf{58.5} & \textbf{74.2} & 100.9G \\ \hline
    \end{tabular}
    \end{center}

\end{table}

\begin{table}[!t]\normalsize
\renewcommand{\arraystretch}{1.0}
\begin{center}
\caption{Ablation study on adding $\textup{C}^2$Former to different stages.}
\centering
\label{table5}
\setlength{\tabcolsep}{1mm}
\begin{tabular}{cccc|c:c}
\hline
\multicolumn{4}{c|}{\textbf{Different Stages with $\textup{C}^2$Former}}              & \multirow{2}{*}{\textbf{mAP}} & \multirow{2}{*}{\textbf{Params}} \\ \cline{1-4}
\textbf{Stage 1} & \textbf{Stage 2} & \textbf{Stage 3} & \textbf{Stage 4} &                                 &                               \\ \hline
                 &                  &                  & \checkmark                & 72.1                 & \textbf{115.7M}                          \\
                 &                  & \checkmark                & \checkmark                & 73.4                          & 129.1M                          \\
                 & \checkmark                & \checkmark                & \checkmark                & \textbf{74.2}                          & 132.5M                 \\
\checkmark                & \checkmark                & \checkmark                & \checkmark                & 74.1                          & 133.3M                         \\ \hline
\end{tabular}
\end{center}
\end{table}

\begin{table}[!t]\normalsize
\caption{Computational cost comparison between $\textup{C}^2$Former and SOTA method.}
\renewcommand{\arraystretch}{1.0}
\setlength{\tabcolsep}{2.5mm}
\begin{center}
\begin{tabular}{ccccccc}
\hline
\textbf{Methods}        & \textbf{mAP}  & \textbf{FLOPs}  & \textbf{Params} \\ \hline
TSFADet    & 73.1         & 109.8G  & 104.7M \\
Ours   & \textbf{73.8}         & \textbf{89.9G}    & \textbf{100.8M}  \\\hline
\end{tabular}
\end{center}
\label{table7}
\end{table}

\begin{figure}[!t]
	\begin{center}
	\includegraphics[scale=0.65]{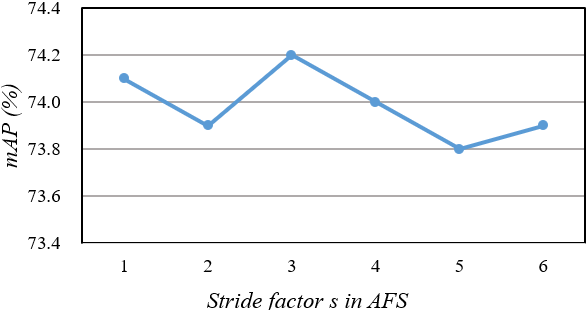}
	\end{center}
    \caption{Ablation study on different downsampled strides $s$.}
    \label{fig7}
\end{figure}

\section{Experiments} \label{experiments}
In this section, we first describe the datasets and evaluation metrics used in our experiments. Then, an exhaustive investigation of ablation studies is presented. Next, we provide the computational cost and visualize the intermediate results of our method. Finally, the comparisons with several closely related detectors are provided.

\subsection{Datasets and Evaluation Metric}
We perform experiments on two public RGB-IR object detection datasets: (1) DroneVehicle Dataset \cite{sun2022drone} and (2) KAIST Multispectral Pedestrian Detection \cite{hwang2015multispectral} (denoted as KAIST dataset). The two datasets contain the pair of RGB-IR images/annotations to perform experiments. 

\noindent\textbf{DroneVehicle dataset.} The DroneVehicle dataset is a large-scale drone-based RGB-Infrared dataset \cite{sun2022drone} that contains 953,087 vehicle instance in 56,878 images. The dataset covers multiple scenarios including urban roads, residential areas, parking lots, and other scenarios from day to night. Besides, rich annotations have been made by the authors with oriented bounding boxes for the five categories (car, bus, truck, van and freight car). The dataset is split into a training set and a test set, and our experimental results are inferred on the test set.

\noindent\textbf{KAIST dataset.} The KAIST dataset is another large-scale dataset with RGB-IR image pairs \cite{hwang2015multispectral}. The dataset consists of 95,328 image pairs with a total of 103,128 pedestrian annotations and 1,182 unique pedestrians. The paired images are captured in a driving environment with regular traffic scenes (campus, street and countryside). Since the problematic annotations in original dataset, our model is trained on new annotations \cite{zhang2021weakly} and inferred using the improved annotations \cite{liu2016multispectral}. The entire test set (All) contains 2,252 frames sampled every 20th frame from videos, among which 1,455 images are captured during daytime (Day) and the rest 797 images are during nighttime (Night). 

\noindent\textbf{Evaluation metric.} For DroneVehicle Dataset, we evaluate the detection performance by utilizing the mean average precision (mAP) as evaluation criteria. For mAP, an Intersection over Union (IoU) threshold of 0.5 is used to calculate True Positives (TP) and False Positives (FP). 
As for KAIST Dataset, we adopt miss rate $\mathbf{MR}^{-2}$ averaged over the false positive per image with range of [$10^-2$, $10^0$] to evaluate the pedestrian detection performance. To fully investigate the performance of the detector, we evaluate the performance using ‘All’ condition \cite{hwang2015multispectral} and six all-day subsets including pedestrian distances and occlusion levels.

\subsection{Implementation Details}
For DroneVehicle dataset, we use single-stage oriented object detector---$\rm S^2$A-Net \cite{han2021align}, the size of initial anchors is set to [4, 8, 16] with a single ratio 1.0, while for KAIST dataset, we utilize a two-stage horizontal object detector---Cascade R-CNN \cite{cai2018cascade}, the size of initial anchors is [4, 8, 16, 32] with a single ratio 2.43. For RPN settings in Cascade R-CNN, we select top-2000 proposals from each image to guarantee the recall value.  We select ResNet50 with FPN (ResNet50-FPN) as the backbone network unless otherwise specified. All detectors are trained for 24 epochs on an NVIDIA TITAN V GPU and optimized using stochastic gradient descent (SGD) algorithm with 0.9 momentum and 0.0001 weight decay. During training, we set two images of size $640 \times 512$ per batch and initial learning rate is 0.001. At epochs 16 and 22, the rate of learning is divided by 10. We perform augmentation to images, such as flip, crop, and mosaic. During inference, both detectors are set up the same as in the training phase and the duplicated bounding boxes are removed by conducting the non-maximum suppression (NMS) with 0.3 IoU threshold.
All experiments are conducted based on the modified code library MMRotate \cite{zhou2022mmrotate} and MMDetection \cite{chen2019mmdetection}. 

\begin{table*}[!t]\normalsize
\renewcommand{\arraystretch}{1.0}
\caption{Detection results (mAP, in \%) on DroneVehicle dataset. Note that all detectors locate and classify vehicles with OBB heads. The results of feature addition using the $\textup{S}^2$A-Net detector are also shown. Best results highlighted in \textbf{bold}.}
\begin{center}

\begin{tabular}{c|c|ccccc:c}
\hline
\textbf{Detectors}      & \textbf{Modality}       & \textbf{Car} & \textbf{Truck} & \textbf{Freight-car} & \textbf{Bus} & \textbf{Van} & \textbf{mAP} \\ \hline
Faster R-CNN \cite{ren2015faster}     & \multirow{6}{*}{RGB}    & 79.0         & 49.0           & 37.2                 & 77.0         & 37.0         & 55.9         \\
ReDet \cite{han2021redet}                  &                         & 80.3         & 56.1           & 42.7                 & 80.2         & 44.4         & 60.8         \\
Oriented R-CNN \cite{xie2021oriented}         &                         & 80.1         & 53.8           & 41.6                 & 85.4         & 43.3         & 60.8         \\
RetinaNet \cite{lin2017focal}        &                         & 78.8         & 39.9           & 19.5                 & 67.3         & 24.9         & 46.1         \\
$\textup{R}^3$Det \cite{yang2021r3det}                 &                         & 79.3         & 42.2           & 24.5                 & 76.0         & 28.5         & 50.1         \\
$\textup{S}^2$A-Net \cite{han2021align}                 &                         & 80.0         & 54.2           & 42.2                 & 84.9         & 43.8         & 61.0         \\ \hline
Faster R-CNN \cite{ren2015faster}     & \multirow{6}{*}{IR}     & 89.4         & 53.5           & 48.3                 & 87.0         & 42.6         & 64.2         \\
ReDet \cite{han2021redet}                  &                         & 90.0         & 61.5           & 55.6                 & 89.5         & 46.6         & 68.6         \\
Oriented R-CNN \cite{xie2021oriented}         &                         & 89.8         & 57.4           & 53.1                 & 89.3         & 45.4         & 67.0         \\
RetinaNet \cite{lin2017focal}         &                         & 89.3         & 38.2           & 40.0                 & 79.0         & 32.1         & 55.7         \\
$\textup{R}^3$Det \cite{yang2021r3det}                   &                         & 89.5         & 48.3           & 16.6                 & 87.1         & 39.9         & 62.3         \\
$\textup{S}^2$A-Net \cite{han2021align}                 &                         & 89.9         & 54.5           & 55.8                 & 88.9         & 48.4         & 67.5         \\ \hline
Halfway Fusion \cite{liu2016multispectral}    & \multirow{7}{*}{RGB+IR} & 90.1         & 62.3           & 58.5                 & 89.1         & 49.8         & 70.0         \\
CIAN \cite{zhang2019cross}             &                         & 90.1         & 63.8           & 60.7                 & 89.1         & 50.3         & 70.8         \\
$\textup{S}^2$A-Net \cite{han2021align}                &                         & 90.0         & 64.5           & 61.7                 & 88.2         & 53.2         & 71.5         \\ 
MBNet \cite{zhou2020improving}             &                         & 90.1         & 64.4           & 62.4                 & 88.8         & 53.6         & 71.9         \\
AR-CNN \cite{zhang2021weakly}           &                         & 90.1         & 64.8           & 62.1                 & 89.4         & 51.5         & 71.6         \\
TSFADet \cite{yuan2022translation}                &                         & 89.9         & 67.9           & 63.7                 & 89.8         & 54.0         & 73.1         \\ 
$\textup{C}^2$Former-$\textup{S}^2$ANet (Ours) &                         & \textbf{90.2}         & \textbf{68.3}           & \textbf{64.4}                 & \textbf{89.8}         & \textbf{58.5}         & \textbf{74.2}         \\ \hline
\end{tabular}
\end{center}
\label{table1}
\end{table*}
\begin{figure}[!t]
	\begin{center}
	\includegraphics[scale=0.32]{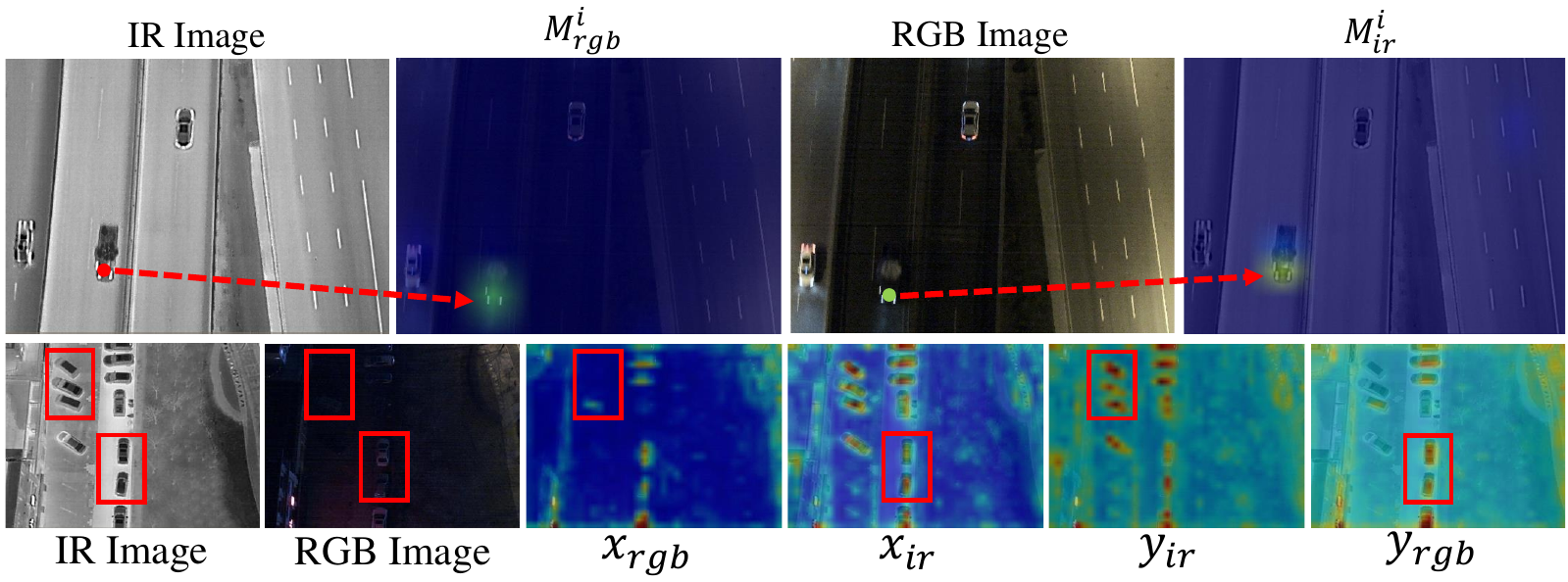}
	\end{center}
    \caption{Visualization of the intermediate results of the ICA module. The first row shows the example of the red and green points corresponding to the attention map in ${M}_{rgb}$ and ${M}_{ir}$, respectively. The second row visualizes the complementary features ${y}_{ir}$ and ${y}_{rgb}$ calculated from the RGB and IR features respectively.}
    \label{fig9}
\end{figure}

\subsection{Ablation Studies}
\noindent\textbf{Ablation on each component.}  We ablate the key components in our $\textup{C}^2$Former to investigate the effectiveness of these designs. The baseline is a two-stream $\rm S^2$A-Net, which only adopts simple additional operation to fuse two modalities features. From Table~\ref{table3}, adopting ICA module can provide +2.3\% significant improvement. The result shows that the ICA module effectively enhances the features of the respective modalities, but it also brings a huge computational cost (FLOPs increase by about 40G). After adding AFS module, the computational complexity is greatly reduced from 139G to 100G, and the model performance is also improved by 0.4\%. Overall, after adding  $\textup{C}^2$Former, mAP increases by 2.7\%, but FLOPs only increases by 3G. 

\noindent\textbf{Ablation on different strides $s$.} 
We also conduct the experiments by modifying the downsampled stride $s$ of AFS module as $s=\{1, 2, 3, 4, 5, 6\}$. As shown in Figure~\ref{fig7}, even we change the stride $s$ of AFS module, performances of our method are still high, which indicates the robustness of $\textup{C}^2$Former to this hyper-parameter. When $s=3$, $\textup{C}^2$Former achieves the highest detection performance.

\noindent\textbf{$\textbf{C}^2$Former at different stages.} 
We add our $\textup{C}^2$Former to the baseline network (two-stream $\rm S^2$A-Net) at different stages. From Table~\ref{table5}, we can observe that after adding  $\textup{C}^2$Former in the last stage, $\rm S^2$A-Net achieves only 72.1\% mAP, while after adding $\textup{C}^2$Former in the second and third stages, it achieves 74.2\% overall accuracy. However, continuing to add $\textup{C}^2$Former to the first stage can not improve detection accuracy but add more parameters.

\subsection{Computational Cost Comparison}
Since $\textup{C}^2$Former is proposed as a plug-in module, for computational cost comparison, we compare with the previous best competitor TSFADet \cite{yuan2022translation}, which also uses a plug-in module called TSRA to perform RGB-infrared object detection. For a fair comparison, we also add $\textup{C}^2$Former to the same baseline network Oriented R-CNN as TSFADet. We test the FLOPs and Params of TSRA module under the same model settings as shown in Table~\ref{table7}. The results show that our detector significantly outperforms TSFADet in all three metrics, which verifies the superiority of our $\textup{C}^2$Former.

\begin{figure}[!t]
    \begin{center}
        \includegraphics[scale=0.25]{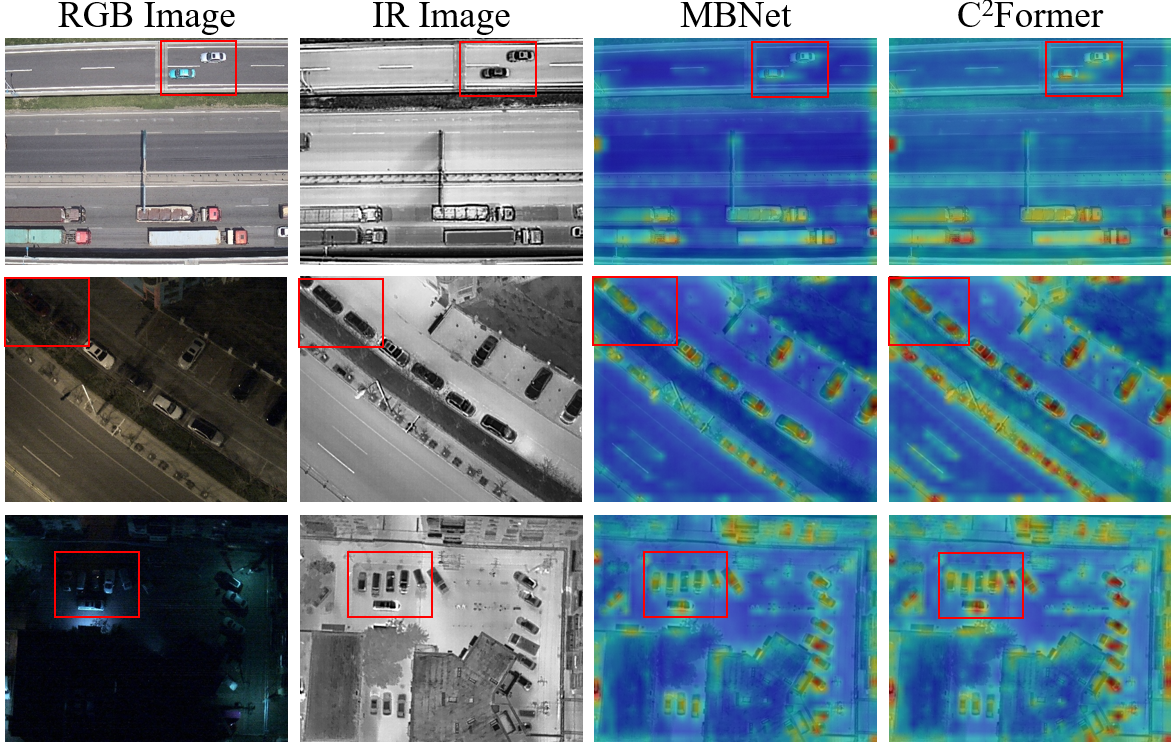}
        \caption{Feature maps visualization comparison of MBNet and our $\textup{C}^2$Former for detection heads. From top to bottom, the RGB-IR image pairs are captured in day, night and dark night, respectively. }
	\label{fig6}
    \end{center}
\end{figure}

\begin{figure}[!t]
    \begin{center}
        \includegraphics[scale=0.32]{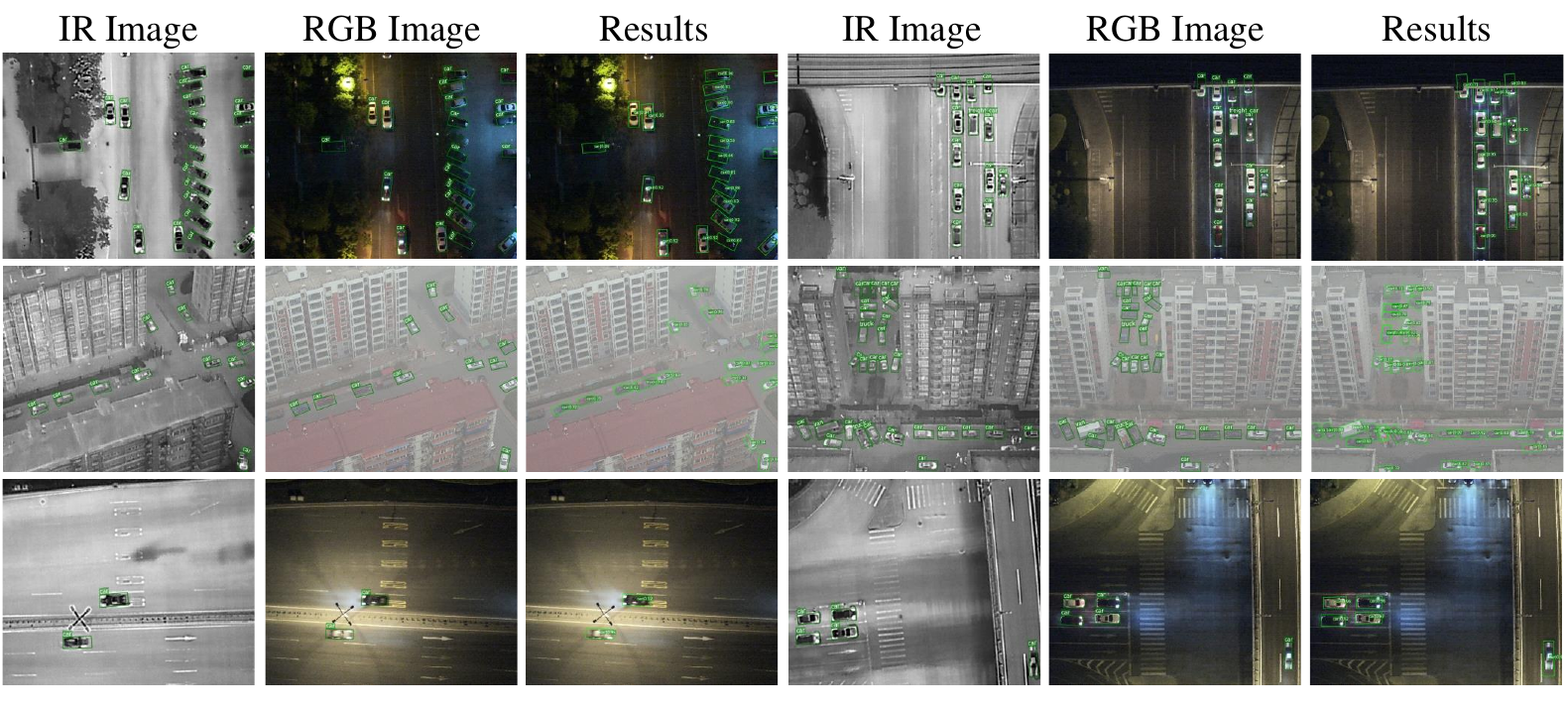}
        \caption{Detection results of our $\textup{C}^2$Former in different scenarios on the DroneVehicle dataset. From top to bottom, objects are in different complex situations of occlusion, low resolution and angle changes.}
	\label{fig13}
    \end{center}
\end{figure}

\begin{figure}[!t]
	\begin{center}
	\includegraphics[scale=0.45]{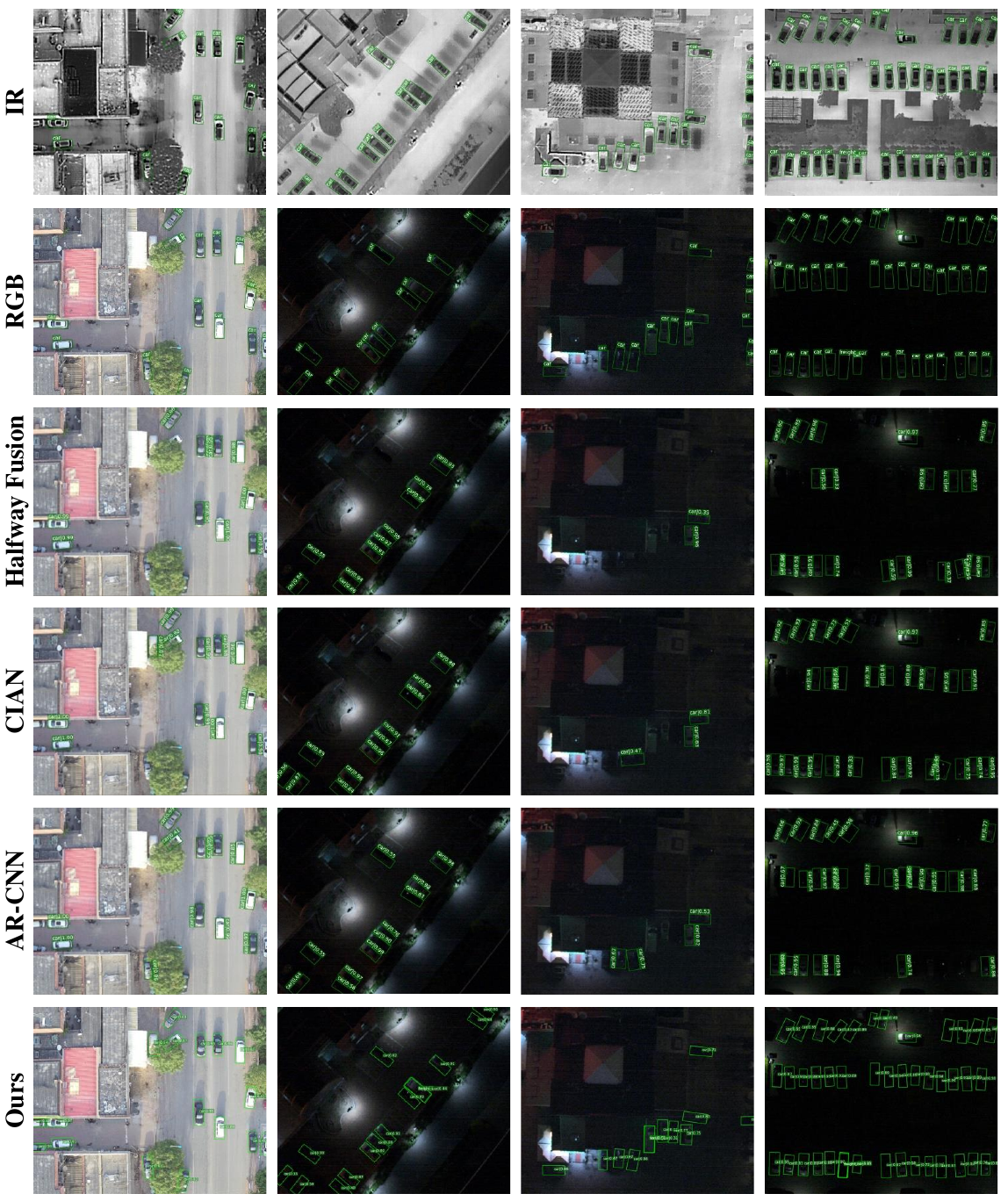}
	\end{center}
    \caption{Visual examples of different methods on the DroneVehicle dataset. From top to bottom are GroundTruth (IR and RGB), Halfway Fusion \cite{liu2016multispectral}, CIAN \cite{zhang2019cross}, AR-CNN \cite{zhang2021weakly} and our $\textup{C}^2$Former-$\textup{S}^2$ANet, respectively. The confidence threshold is set to 0.3 when visualizing these results.}
	\label{fig12}
\end{figure}

\subsection{Visualization Results}
\noindent\textbf{Intermediate results of ICA module.} 
To illustrate the effectiveness of  ICA module, we visualize the intermediate results in Figure~\ref{fig9}. From the first row, we find that the similarity map corresponding to the red and green points can clearly reflect the unaligned position in the other modality, which shows this similarity map indeed has the ability to localize the calibration position. From the $y_{ir}$ and $y_{rgb}$ in the second row, we can observe that the ICA module can dynamically complement the RGB and IR features according to the intensity of the $x_{rgb}$ and $x_{ir}$. For example, the obscure features $x_{rgb}$ and $x_{ir}$ in the red box are complemented. The results show that the ICA module achieves both feature calibration and complementary fusion.

\noindent\textbf{Visualization of feature map.} 
Since the MBNet \cite{zhou2020improving} method is a popular method by fusing RGB-IR features in the backbone, we also visualize the feature maps (shown in Figure~\ref{fig6}) of MBNet method and $\textup{C}^2$Former respectively. We note that vehicles show more contrast with the background in the feature maps generated by $\textup{C}^2$Former for all capturing times. During the daytime (first row), objects with miscalibration problem (inside the red box) have better feature representation by using $\textup{C}^2$Former. Furthermore, when the objects are in the low-light condition (second and third rows), $\textup{C}^2$Former can complement corresponding features with global information to enhance the feature representation compared to the MBNet method.

\noindent\textbf{Detection results in complex scenarios.}
To illustrate the detection capabilities of our model in complex scenarios, we visualize the results of the model in different scenarios. As shown in Figure~\ref{fig13}, the objects in these images are in three complex situations: occlusion, low resolution and angle change. Through these visual results, we find that $\textup{C}^2$Former is robust to the above situations, which verifies its effectiveness.

\begin{table*}[!t]\normalsize
\renewcommand{\arraystretch}{1.0}
\caption{Detection results (MR, in\%) under the IoU threshold (\textbf{0.5}) of different pedestrian distances, occlusion levels and light conditions (Day and Night) on KAIST dataset. The pedestrian distances consist of 'Near' (115 $\leq$ \emph{height}), 'Medium' (45 $\leq$ \emph{height} $<$ 115) and 'Far' (1 $\leq$ \emph{height} $<$ 45), while occlusion levels consist of 'None' (never occluded), 'Partial' (occluded to some extent up to one half) and 'Heavy' (mostly occluded). The results of feature addition using the Cascade R-CNN detector are also shown. The number in \textbf{bold} indicates the best score.}
\begin{center}
\begin{tabular}{c|cccccc:cc:c}
\hline
\textbf{Methods}          & \textbf{Near}          & \textbf{Medium}         & \textbf{Far}            & \textbf{None}           & \textbf{Partial}        & \textbf{Heavy}          & \textbf{Day}            & \textbf{Night}          & \textbf{All}            \\ \hline
ACF \cite{wagner2016multispectral}             & 28.74         & 53.67          & 88.20          & 62.94          & 81.40          & 88.08          & 64.31          & 75.06          & 67.74          \\
Halfway Fusion \cite{liu2016multispectral}  & 8.13          & 30.34          & 75.70          & 43.13          & 65.21          & 74.36          & 47.58          & 52.35          & 49.18          \\
FusionRPN+BF \cite{konig2017fully}     & 0.04          & 30.87          & 88.86          & 47.45          & 56.10          & 72.20          & 52.33          & 51.09          & 51.70          \\
IAF R-CNN \cite{li2019illumination}       & 0.96          & 25.54          & 77.84          & 40.17          & 48.40          & 69.76          & 42.46          & 47.70          & 44.23          \\
IATDNN+IASS \cite{guan2019fusion}      & 0.04          & 28.55          & 83.42          & 45.43          & 46.25          & 64.57          & 49.02          & 49.37          & 48.96          \\
CIAN \cite{zhang2019cross}            & 3.71          & 19.04          & 55.82          & 30.31          & 41.57          & 62.48          & 36.02          & 32.38          & 35.53          \\
MSDS-RCNN \cite{li2018multispectral}       & 1.29          & 16.19          & 63.73          & 29.86          & 38.71          & 63.37          & 32.06          & 38.83          & 34.15          \\
AR-CNN \cite{zhang2021weakly}          & 0.00          & 16.08          & 69.00          & 31.40          & 38.63          & \textbf{55.73} & 34.36          & 36.12          & 34.95          \\
MBNet \cite{zhou2020improving}           & 0.00          & 16.07          & 55.99          & 27.74          & 35.43          & 59.14          & 32.37          & 30.95          & 31.87          \\ \hline
Cascade R-CNN \cite{cai2018cascade} & 2.23 & 19.73 & 52.21 & 28.43 & 41.05 & 63.89          & 35.29 & 28.57 & 33.35 \\
$\textup{C}^2$Former-CascadeRCNN (Ours) & \textbf{0.00} & \textbf{13.71} & \textbf{48.14} & \textbf{23.91} & \textbf{32.84} & 57.81          & \textbf{28.48} & \textbf{26.67} & \textbf{28.39} \\ \hline
\end{tabular}
\end{center}
\label{table2}
\end{table*}

\begin{table*}[!t]\normalsize
\renewcommand{\arraystretch}{1.0}
\caption{Detection results (MR, in\%) under the IoU threshold (\textbf{0.75}) of different pedestrian distances, occlusion levels and light conditions (Day and Night) on KAIST dataset. The number in \textbf{bold} indicates the best score.}
\label{table6}
\centering
\begin{tabular}{c|cccccc:cc:c}
\hline
\textbf{Methods}          & \textbf{Near}          & \textbf{Medium}         & \textbf{Far}            & \textbf{None}           & \textbf{Partial}        & \textbf{Heavy}          & \textbf{Day}            & \textbf{Night}          & \textbf{All}            \\ \hline
ACF \cite{wagner2016multispectral}             & 88.53         & 89.68          & 99.18          & 92.63          & 96.47          & 98.91          & 93.28          & 95.01          & 93.73          \\
Halfway Fusion \cite{liu2016multispectral}  & 68.71          & 85.56          & 98.84           & 88.60          & 95.27          & 97.89          & 89.16          & 92.93          & 90.38          \\
FusionRPN+BF \cite{konig2017fully}     & 56.16          & 80.42          & 99.92          & 84.96          & 91.69          & 95.84          & 84.84          & 91.14          & 86.98          \\
IAF R-CNN \cite{li2019illumination}       & 57.39          & 82.68          & 99.32          & 86.27          & 91.12          & 97.14          & 86.67          & 90.85          & 88.07          \\
IATDNN+IASS \cite{guan2019fusion}      & 64.12          & 83.23          & 99.74          & 87.21          & 91.15          & 96.69          & 88.98          & 88.91          & 88.92          \\
CIAN \cite{zhang2019cross}            & 58.80          & 78.16          & 93.45          & 82.74          & 89.39          & 96.29          & 83.56          & 88.66          & 85.28          \\
MSDS-RCNN \cite{li2018multispectral}       & 62.73          & 73.73          & 97.23          & 81.00          & 86.27          & 95.27          & 81.81          & 88.80          & 83.43          \\
AR-CNN \cite{zhang2021weakly}          & 44.95          & 70.87          & 95.99          & 77.45          & 84.74          & 93.42              & 76.98          & 87.34          & 80.19          \\
MBNet \cite{zhou2020improving}           & \textbf{36.49}          & 67.39          & 92.45          & 74.41          & 81.33          & 91.46          & 74.86          & 82.11          & 77.53          \\ \hline
Cascade R-CNN \cite{cai2018cascade} & 51.89 & 76.53 & 93.38 & 76.82 & 86.46 & 93.13  & 77.54 & 80.78 & 78.57 \\
$\textup{C}^2$Former-CascadeRCNN (Ours) & 38.58 & \textbf{65.20} & \textbf{87.62} & \textbf{72.02} & \textbf{79.14} & \textbf{91.37}          & \textbf{74.18} & \textbf{79.18} & \textbf{75.50} \\ \hline
\end{tabular}
\end{table*}

\subsection{Evaluation on DroneVehicle Dataset}
As for RGB-IR vehicle detection task, we inject our proposed $\textup{C}^2$Former into $\rm S^2$A-Net \cite{han2021align} to construct a new oriented object detector called \emph{${C}^2$Former-$S^2$A-Net}, and compare it with 6 state-of-the-art single-modality detectors, including Faster R-CNN \cite{ren2015faster}, ReDet \cite{han2021redet}, Oriented R-CNN \cite{xie2021oriented}, RetinaNet \cite{lin2017focal}, $\rm R^3$Det \cite{yang2021r3det} and $\rm S^2$A-Net \cite{han2021align}. We also re-implement five RGB+IR multispectral methods (Halfway Fusion \cite{liu2016multispectral}, CIAN \cite{zhang2019cross}, MBNet \cite{zhou2020improving}), AR-CNN \cite{zhang2021weakly} and TSFADet \cite{yuan2022translation}) for RGB-IR object detection on rotation detectors. Table~\ref{table1} shows the results of detection accuracy. In single-modality methods, the detection performance on IR modality is significantly better than RGB modality. Through the results, Oriented R-CNN and $\rm S^2$A-Net have comparable detection accuracy (67.0\% mAP and 67.5\% mAP) in two-stage and single-stage object detectors, respectively. Obviously, the multispectral detection methods using both RGB and IR modalities are superior to the single-modality methods, and our $\textup{C}^2$Former-$\rm S^2$ANet achieves 74.2\% mAP, which is the highest detection accuracy among these multispectral detection methods.

For DroneVehicle dataset, we also provide some visual detection results of the compared methods on the validation dataset in Figure~\ref{fig12}. In comparison, since the paired vehicle objects exist modality miscalibration or fusion imprecision problem,  some objects are misidentified or not detected in Halfway Fusion \cite{liu2016multispectral}, CIAN \cite{zhang2019cross} and AR-CNN \cite{zhang2021weakly} methods. These results demonstrate that the ICA module can provide aligned and complementary features, enabling our detector to achieve better classification and localization processes.

\begin{figure*}[!t]
	\begin{center}
	\includegraphics[scale=0.135]{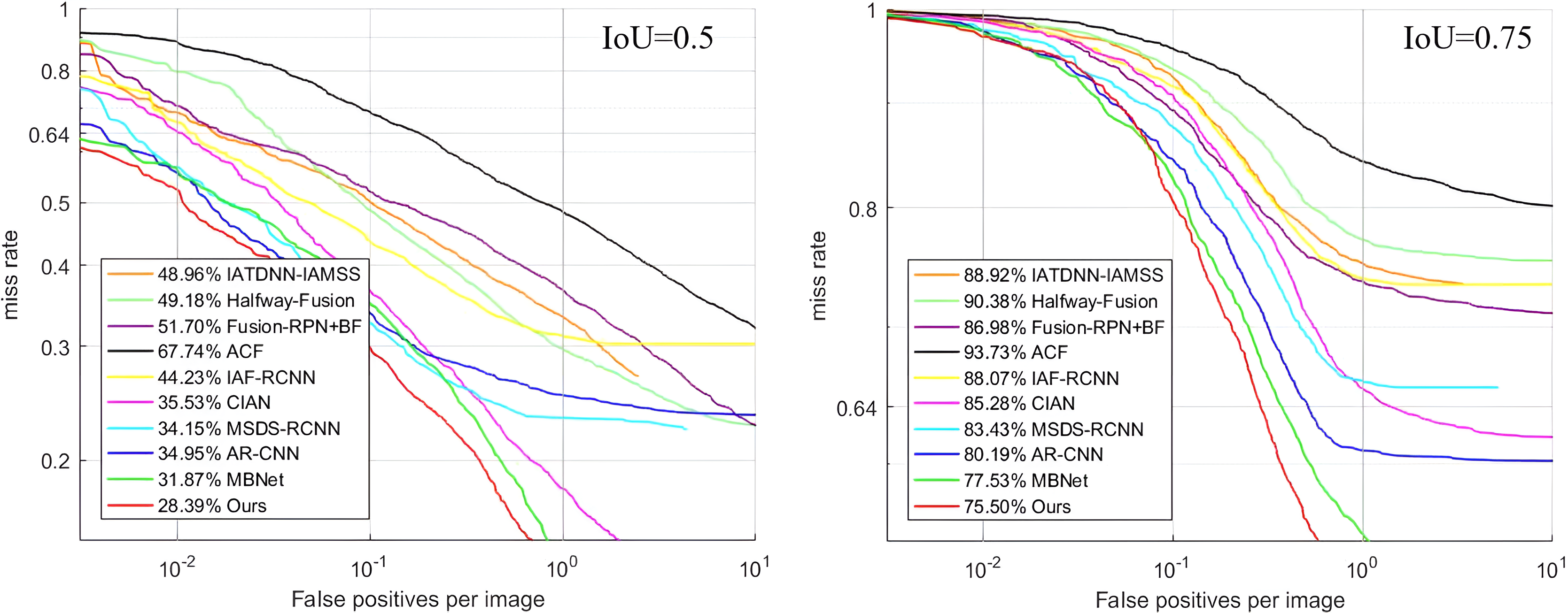}
	\end{center}
    \caption{Miss rate plotted against False Positives Per Image (FPPI) under the IoU threshold of 0.5 (left) and 0.75 (right) for KAIST dataset.}
	\label{fig10}
\end{figure*}

\begin{figure}[!t]
	\begin{center}
	\includegraphics[scale=0.58]{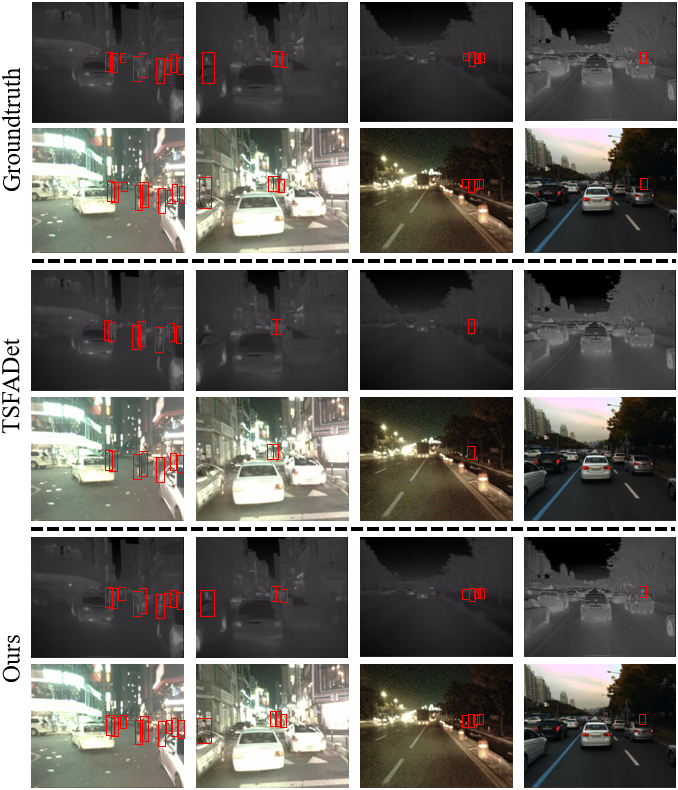}
	\end{center}
    \caption{Visual examples of RGB and Infrared detection on the KAIST dataset for our method and TSFADet \cite{yuan2022translation}, respectively. }
	\label{fig11}
\end{figure}

\subsection{Evaluation on KAIST Dataset}
For multispectral pedestrian detection task, we build a horizontal detector specific to pedestrian detection called \emph{${C}^2$Former-CascadeRCNN} by implementing our $\textup{C}^2$Former on Cascade R-CNN \cite{cai2018cascade}. In the experiments, our proposed detector is compared with several state-of-the-art multispectral pedestrian algorithms, including ACF \cite{wagner2016multispectral}, Halfway Fusion \cite{liu2016multispectral}, FusionRPN+BF \cite{konig2017fully}, IAF R-CNN \cite{li2019illumination}, IATDNN+IASS \cite{guan2019fusion}, CIAN \cite{zhang2019cross}, MSDS-RCNN \cite{li2018multispectral}, AR-CNN \cite{zhang2021weakly}. Table~\ref{table2} shows our proposed detector obtain 28.48 MR, 26.67 MR, and 28.39 MR on the 'Day', 'Night' and 'All' subset respectively under the IoU threshold of 0.5, all these scores are lower than the previous best competitor MBNet \cite{zhou2020improving}. We also show the detection results under the IoU threshold of 0.75 on KAIST dataset in Table~\ref{table6}. In the case of a stricter IoU threshold of 0.75, we can observe that the performance of all detectors has a significant drop. Our proposed method achieves about 2\% lower than MBNet \cite{zhou2020improving} which implies that the $\textup{C}^2$Former-Cascade R-CNN has a better localization accuracy compared with other methods. The larger the IoU threshold is set, the harder the predicted bounding boxes are considered to be True Positives (TP) in the evaluation. 
We also test our detectors under six subsets in terms of pedestrian distance and occlusion level also shown in Table~\ref{table2}. The results show that $\textup{C}^2$Former-Cascade R-CNN has the better performance than other multispectral approaches with no specific technique to handle the small and occlusion pedestrians. Especially, $\textup{C}^2$Former-Cascade R-CNN achieves about 8\% and 5\% lower than MBNet on the 'Far' subset under the IoU threshold of 0.5 and 0.75 respectively,  which indicates the detector with $\textup{C}^2$Former can be more effective for small object detection. 

To display the experimental results more intuitively, we also plot the MR curves under the different IoU thresholds as shown in Figure~\ref{fig10} and provide some visual detection results of the compared methods in Figure~\ref{fig11}. In comparison, since the paired vehicle objects exist modality miscalibration or fusion imprecision problem,  some objects are misidentified or not detected in Halfway Fusion \cite{liu2016multispectral}, CIAN \cite{zhang2019cross} and AR-CNN \cite{zhang2021weakly} methods. 
All these results demonstrate that the ICA module can provide aligned and complementary features, enabling our detector to achieve better classification and localization processes. All the results demonstrate that $\textup{C}^2$Former can fully utilize the RGB-IR complementary information and achieve a high localization accuracy.

\section{Conclusion} \label{conclusion}
In this work, we proposed a novel calibrated and complementary transformer for RGB-IR object detection called $\textup{C}^2$Former, which can effectively tackle with miscalibration and fusion imprecision problems. The proposed $\textup{C}^2$Former mainly consisted of two modules: an Inter-modality Cross-Attention (ICA) module to achieve complementary fusion and an Adaptive Feature Sampling (AFS) module to reduce computational complexity. The proposed $\textup{C}^2$Former had good compatibility and can be injected into existed object detector frameworks. Specifically, We constructed $\textup{C}^2$Former-$\textup{S}^2$ANet and $\textup{C}^2$Former-CascadeRCNN and conducted extensive experiments on DroneVehicle and KAIST detection datasets. The results demonstrated the superiority of our $\textup{C}^2$Former-based detectors. We believe that our method can be applied to various studies in the RGB-IR object detection areas.

\bibliographystyle{IEEEtran}
\bibliography{egbib}

\end{document}